%% file: main.tex
\documentclass[a4paper, conference]{IEEEtran}
\IEEEoverridecommandlockouts
\usepackage{cite}
\usepackage{amsmath,amssymb,amsfonts}
\usepackage{graphicx}
\usepackage{textcomp}
\usepackage{xcolor}

\usepackage[symbol]{footmisc}

\include{definitions}

\def\BibTeX{{\rm B\kern-.05em{\sc i\kern-.025em b}\kern-.08em
    T\kern-.1667em\lower.7ex\hbox{E}\kern-.125emX}}
\begin{document}

\title{DAIL: Dataset-Aware and Invariant Learning for Face Recognition$^*$}

\author{\IEEEauthorblockN{Gaoang Wang$^{1,2}$, Lin Chen$^{1}$, Tianqiang Liu$^{1}$, Mingwei He$^{1}$, and Jiebo Luo$^{3}$}
\IEEEauthorblockA{$^1$\textit{Wyze Labs, Kirkland, WA 98033, USA}\\
$^2$\textit{Zhejiang University-University of Illinois at Urbana-Champaign Institute, Haining, Zhejiang 314400, China}\\
$^3$\textit{University of Rochester, Rochester, NY 14627, USA}\\
Email: gaoangwang@intl.zju.edu.cn, $\{$lchen, tliu, mhe$\}$@wyze.com, jluo@cs.rochester.edu}}


\maketitle

\begin{abstract}
To achieve good performance in face recognition, a large scale training dataset is usually required. A simple yet effective way to improve the recognition performance is to use a dataset as large as possible by combining multiple datasets in the training. However, it is problematic and troublesome to naively combine different datasets due to two major issues. First, the same person can possibly appear in different datasets, leading to an identity overlapping issue between different datasets. Naively treating the same person as different classes in different datasets during training will affect back-propagation and generate non-representative embeddings. On the other hand, manually cleaning labels may take formidable human efforts, especially when there are millions of images and thousands of identities. Second, different datasets are collected in different situations and thus will lead to different domain distributions. Naively combining datasets will make it difficult to learn domain invariant embeddings across different datasets.
In this paper, we propose DAIL: Dataset-Aware and Invariant Learning to resolve the above-mentioned issues. To solve the first issue of identity overlapping, we propose a dataset-aware loss for multi-dataset training by reducing the penalty when the same person appears in multiple datasets. This can be readily achieved with a modified softmax loss with a dataset-aware term. To solve the second issue, domain adaptation with gradient reversal layers is employed for dataset invariant learning. The proposed approach not only achieves the state-of-the-art results on several commonly used face recognition validation sets, including LFW, CFP-FP, and AgeDB-30, but also shows great benefit for practical use.    
\end{abstract}


\footnotetext[1]{This work was conducted while the first author had the full-time position at Wyze Labs.}

\section{Introduction}
Face recognition has received much attention in recent years and has been widely used in many industrial fields, such as security, surveillance and mobile applications. Many deep learning based state-of-the-art methods \cite{deng2019arcface, duan2019uniformface, kang2019attentional, liu2017sphereface, wang2018cosface, wen2016discriminative, zhao2019regularface} are introduced and more and more accurate models have been achieved.

Many existing state-of-the-art (SOTA) methods explore different loss functions to achieve good performance in face recognition. Most of the loss functions aim at learning face embeddings that maximize the inter-class distance and minimize the intra-class distance. Typically, two types of losses are commonly used. One is softmax-based classification loss with several variations, such as SphereFace \cite{liu2017sphereface, liu2016large}, CosFace \cite{wang2018cosface, wang2018additive}, and ArcFace \cite{deng2019arcface}. The other is the contrastive loss, including triplet loss \cite{schroff2015facenet}, center loss \cite{wen2016discriminative}, range loss \cite{zhang2017range} and margin loss \cite{deng2017marginal}. 

Embedding network architecture is another important factor in face recognition. Some commonly used feature extractors include VGG \cite{parkhi2015deep}, ResNet \cite{he2016deep}. Some modifications like squeeze-and-excitation (SE) \cite{hu2018squeeze} module, group convolutions \cite{krizhevsky2012imagenet}, can be also applied to the existing backbones. Generally speaking, for the same type of architectures, larger models tend to have better performance. There are also some light-weighted backbones, like MobileNet \cite{howard2017mobilenets, sandler2018mobilenetv2, howard2019searching}, EfficientNet \cite{tan2019efficientnet}, which enable the face recognition to run on mobile devices and cameras. This type of carefully designed architectures only has a marginal impact on the accuracy drop but saves a lot of computational cost. 

\begin{figure}[!t]
\centerline{\includegraphics[width=\linewidth]{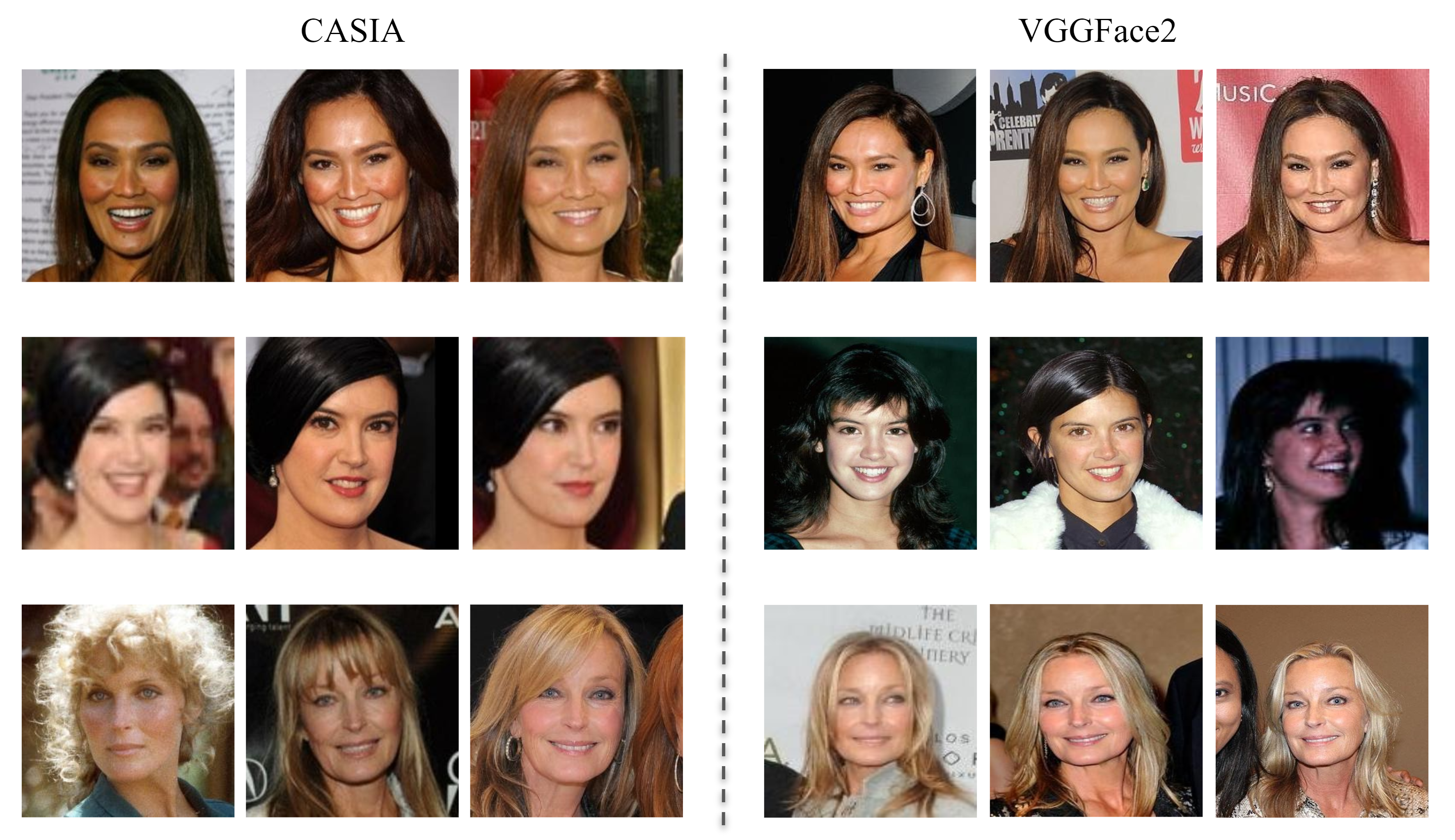}}
\caption{Examples of the ID overlapping issue across different datasets. Each row represents the faces of the same person. The left and right parts of the figure represent the images selected from CASIA and VGGFace2, respectively.}
\label{fig:id_overlap}
\end{figure}

\begin{figure}[!t]
\centerline{\includegraphics[width=\linewidth]{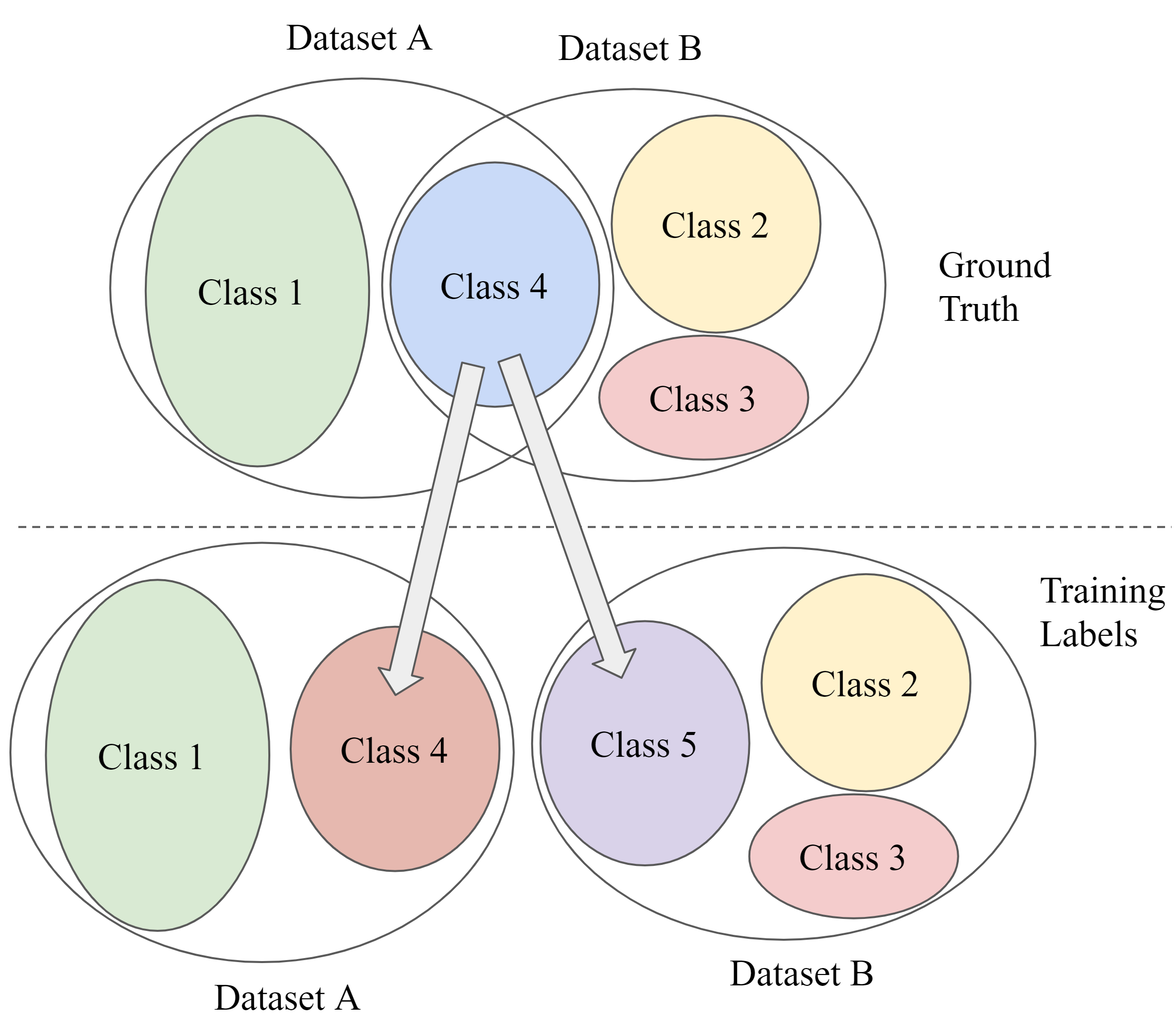}}
\caption{Illustration of the ID overlapping issue across datasets. For example, class 4 exists in both datasets A and B. When combining different datasets in  training, if we naively set distinct labels and exclude each other across different datasets, faces from class 4 will be set to two different labels in the training, leading to the ID overlapping issue, which will negatively affect the recognition performance.}
\label{fig:issue}
\end{figure}

\begin{figure*}[!t]
\begin{center}
\includegraphics[width=\linewidth]{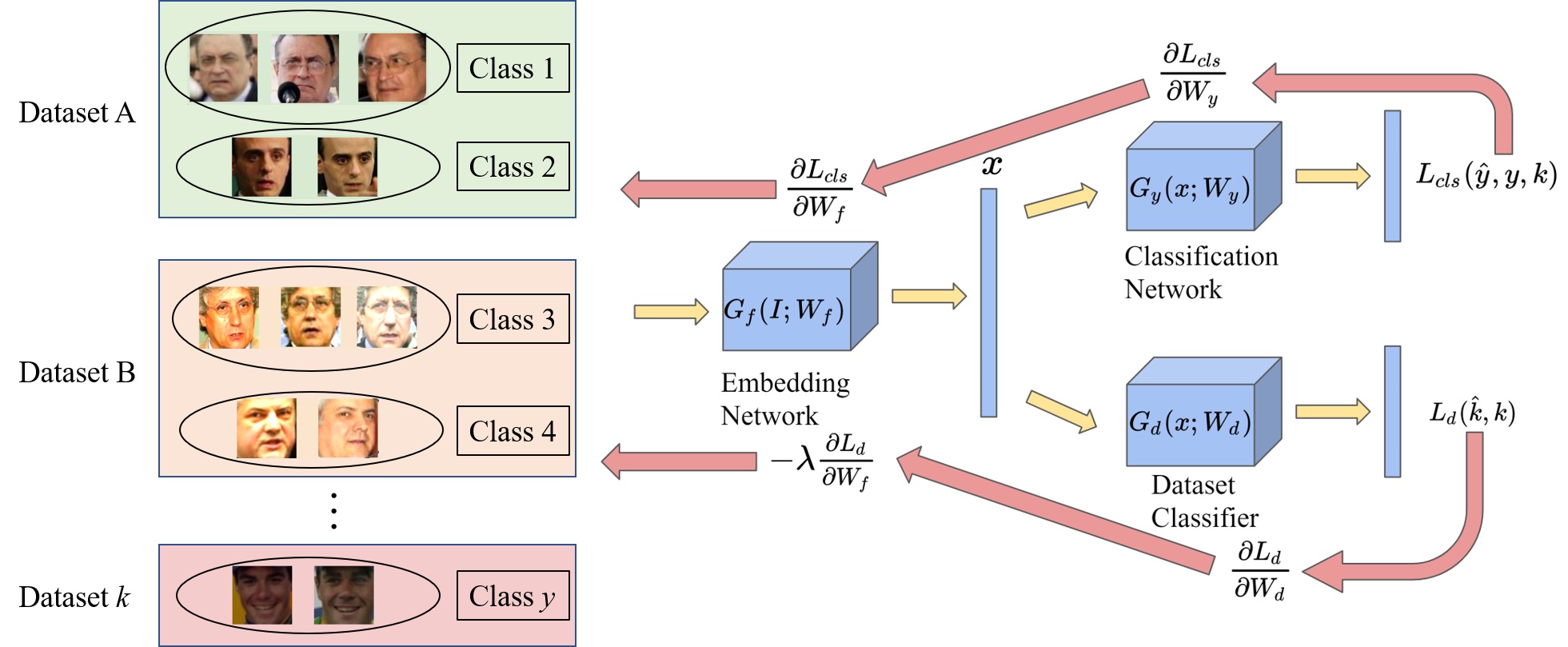}
\end{center}
   \caption{The framework of the proposed method. On the left side of the figure, we show that multiple datasets (displayed in different colors) are combined in  training and each ellipse represents a distinct face ID. The model contains three sub-networks, \ie, the embedding network $G_f$, the classification network $G_y$, and dataset classifier $G_d$. Yellow arrows represent the forward flow while the red arrows represent the back-propagation for the gradient update. Note that the dataset-aware loss is supervised by both class label $y$ and the dataset index $k$.}
\label{fig:flowchart}
\end{figure*}

Training data is also very critical for learning a good model. A few years ago, DeepID \cite{sun2014deep} includes only a few hundreds of thousands of images in the training. Recently, more and more large scale datasets have been created, such as VGGFace2 \cite{cao2018vggface2}, MS1M \cite{guo2016ms}, MegaFace \cite{kemelmacher2016megaface}, CASIA \cite{yi2014learning}. FaceNet \cite{schroff2015facenet} even takes over 200 million images in the training. With the help of these large scale datasets, significant progress has been made for face recognition in recent years. 

A simple and straight forward idea to achieve a further improvement is to combine all the available face recognition datasets in the training. However, this idea is rarely explored or mentioned in the literature. The main challenge of dataset fusion is the label cleaning issue. As we know, some existing datasets contain a lot of public celebrity images. ID overlapping across different datasets is a very common issue in face recognition, \ie, several datasets may contain the same person or face images. Take CASIA \cite{yi2014learning} and VGGFace2 \cite{cao2018vggface2} for example. There are a lot of the overlapping identities, as shown in Fig.~\ref{fig:id_overlap}. In the training, if we naively set distinct class IDs for images from different datasets even if they are actually from the same person, the incorrect labels will harm the classification for multi-dataset training. This issue is illustrated in Fig.~\ref{fig:issue} for better understanding. To overcome such issues, label cleaning is one approach that carefully checks the face ID and images across different datasets, but this may take a lot of computational cost and human effort. Whenever a new face recognition dataset is being created, there would be a huge trouble and effort to check whether certain face images have been already included in the existing datasets, in order to avoid the ID overlapping issue across datasets.

To address the challenge of combing multiple datasets training in face recognition, we propose a novel approach with a designed dataset-aware loss, aiming at large scale multi-dataset training without any prerequisites for label cleaning. To be specific, the dataset-aware loss is built upon the commonly used softmax loss with a binary dataset indicator. Whenever a training sample comes, the softmax is calculated within the dataset it belongs to with no influence from other datasets. As a result, the ID overlapping issue is largely alleviated. Meanwhile, to further ensure the face embeddings are not dataset dependent, dataset invariant learning with gradient reversal layers (GRL) \cite{ganin2014unsupervised} is adopted for multi-dataset domain adaptation. The entire training framework is illustrated in Fig.~\ref{fig:flowchart}. 

We summarize the contributions as follows:
\begin{itemize}
\item We investigate multi-dataset training in face recognition where the same identity can appear in multiple datasets, leading to the ID overlapping issue, and design a dataset-aware loss to solve the ID overlapping issue without any effort on label cleaning and unsupervised soft label approaches. 
\item We applied domain adaptation with gradient reversal layers to ensure the robustness of multi-dataset training and to learn dataset invariant face embeddings.
\item We conduct comprehensive experiments to show the effectiveness of each component of the proposed method, which has achieved the state-of-the-art results on face verification tasks.
\end{itemize}

\section{Related Work}

\subsection{Loss Function}
\textbf{Classification Loss.} Classification-based losses, including softmax loss and its variations, are widely used in face recognition \cite{liu2017sphereface, liu2016large, wang2018cosface, wang2018additive, deng2019arcface, chen2017noisy, wan2018rethinking, qi2018face}. In recent years, angular margin based softmax loss shows great power in face recognition, as discussed in \cite{liu2017sphereface, liu2016large, wang2018cosface, wang2018additive, deng2019arcface}. For example, SphereFace \cite{liu2017sphereface, liu2016large} introduces an angular margin to the softmax; CosFace \cite{wang2018cosface, wang2018additive} proposes a large margin cosine loss where an extra margin is applied in cosine space rather than the angle space, and ArcFace \cite{deng2019arcface} employs an additive angular margin to the softmax. With such margin modifications, the feature embeddings are more discriminative across different IDs. 

Rather than using angular based margin softmax, other variations are also explored in \cite{chen2017noisy, wan2018rethinking, qi2018face}.
In \cite{chen2017noisy}, noisy softmax is proposed to mitigate the early saturation issue by injecting annealed noise in softmax. Gaussian mixture loss is introduced with the assumption that the deep features follow the Gaussian mixture distribution in \cite{wan2018rethinking}. Moreover, \cite{qi2018face} proposes a centralized coordinate learning approach with angular margin enhancement. 

\textbf{Contrastive Loss.} Contrastive loss is commonly used in the distance metric learning field and is also well explored in face recognition. This type of loss aims at learning face embeddings that maximize the inter-class distance while minimize the intra-class distance within the batch samples. Examples include \cite{schroff2015facenet, wen2016discriminative, zhang2017range, deng2017marginal}. FaceNet \cite{schroff2015facenet} first introduces triplet loss that encourages the distance learning from an anchor sample to the positive and negative samples in a triplet manner. Then center loss is proposed in \cite{wen2016discriminative} to penalize the discrepancy between the deep features and their corresponding class centers. Besides that, range loss \cite{zhang2017range} is designed to reduce overall intra-personal variations while enlarging inter-personal differences simultaneously. Similarly, the marginal loss \cite{deng2017marginal} simultaneously minimizes the intra-class variances and maximizes the inter-class distances by focusing on the marginal samples.

Since the focus of classification loss and contrastive loss has a small difference, combining these two types of losses is also commonly used in the training, such as \cite{wen2016discriminative, deng2017marginal}. Despite their benefit for generating discriminative features, none of these proposed losses can address the mislabeling issues for multi-dataset training.

\subsection{Handling Noisy Data}
\textbf{Label Cleaning.} The noisy label is a common issue in face recognition. Several approaches are proposed for label cleaning \cite{jin2018community, ng2014data, varkarakis2020dataset, wang2018devil}, aiming at generating a clean dataset from noisy annotated labels. For example, a comprehensive study of noisy data is summarized and data cleaning approaches are investigated in \cite{wang2018devil}. A graph-based cleaning method that employs the community detection algorithm and deep CNN models to delete mislabeled images is proposed in \cite{jin2018community}. Besides that, identifying and removing the wrong labeled face images is formulated as a quadratic programming problem in \cite{ng2014data}. Furthermore, with a simpler strategy \cite{varkarakis2020dataset}, pre-trained face recognition models are applied directly for label cleaning. However, such automatic or semi-automatic approaches usually have very high computational cost. How to efficiently remove overlapping labels from different datasets are seldom touched.

\textbf{Noise-Resistant Learning.} Rather than cleaning the dataset, there are also many noise-resistant approaches \cite{wu2018light, yang2019feature, hu2019noise, wang2019co} that can alleviate the effect from noisy data in the training. For example, \cite{wu2018light} designs a light CNN for face representation with noisy labels. Besides that, a data filtering method is proposed in \cite{yang2019feature} to automatically filter out the data with incorrect labels in the training stage. Furthermore, as presented in \cite{hu2019noise, wang2019co}, sample weighting strategies are well explored and empirically proved to be also effective ways for handling noisy labels. These noise-resistant online learning approache do not require any cleaning step in advance, and thus can save much computational cost. However, the error can be easily propagated in the long time training with pseudo corrected labels and the ID overlapping issue is still not well addressed for multi-dataset training. 

\subsection{Domain Adaptation}
Domain adaptation \cite{wu2019disentangled, song2018adversarial, wen2018improving, de2018heterogeneous, he2019cross, banerjee2016soft, luo2018deep, kan2014domain, hong2017sspp, crosswhite2018template, liu2016transferring, wang2019racial} also plays a very important role in face recognition to deal with the domain drift issue between training datasets and testing datasets.
Transfer learning is one of the most straightforward approaches for domain adaptation \cite{wen2018improving, crosswhite2018template, liu2016transferring}. For example, fine-tuning approaches are explored in \cite{wen2018improving}. Template adaptation is used in \cite{crosswhite2018template} for face verification and identification. Transfer learning with triplet loss \cite{liu2016transferring} is employed for bridging the gap between different domains. However, transfer learning based approaches cannot be easily applied to unlabeled target domain images.

Domain specific architecture design also shows effectiveness in face recognition \cite{de2018heterogeneous}. Specifically, in \cite{de2018heterogeneous}, a domain speciﬁc unit architecture is proposed for each domain, aiming at extracting different low-level features from different domains. However, such methods require several sub-networks for each domain and are not efficient for practical usage.

Besides that, directly transferring the face images from the source to the target domain is also one commonly used approach for domain adaptation \cite{song2018adversarial, he2019cross, kan2014domain, hong2017sspp}. For example, an image generator is applied to transform the image from the source domain to the target domain in \cite{song2018adversarial, he2019cross}. Using a linear combination of sparse target domain neighbors in the image space to represent the source images is proposed in \cite{kan2014domain}. In \cite{hong2017sspp}, a generative approach with the help of a 3D face model is investigated for a single sample face recognition. Such generative approaches show great power in the domain adaptation field but usually require a large dataset for training.

Alternatively,  domain adaptation can also be  conducted in the latent feature space \cite{wu2019disentangled, banerjee2016soft, luo2018deep, wang2019racial}. For example, disentangled variational representation is proposed in \cite{wu2019disentangled} for cross-model matching. In \cite{banerjee2016soft}, a simple SVM-like model is applied to transform the latent feature space for the adaptation. Maximum mean discrepancy based approaches are also proved to be effective in \cite{luo2018deep, wang2019racial}. Compared with the direct methods that generate target images directly, such latent space adaptation methods are more efficient and robust. However, combining multiple datasets in the training for the adaptation is rarely explored.

\section{Proposed Method}

To achieve a better performance, combining different datasets in training is a straightforward strategy. As we know, ID overlapping across different datasets is a very common issue in face recognition. In the training, the same face ID from different datasets is treated as different labels. Such mislabeled examples will largely affect the recognition performance. Label cleaning is one approach to overcome such issues, but it requires a lot of human effort. 

In the following subsections, we present the dataset-aware loss that can be utilized in the multi-dataset training without any label cleaning effort. The dataset-aware loss can be easily combined with existing state-of-the-art softmax based losses, like SphereFace \cite{liu2017sphereface, liu2016large}, CosFace \cite{wang2018cosface, wang2018additive} and ArcFace \cite{deng2019arcface}. Meanwhile, we also employ the domain adaptation approach with gradient reversal layers (GRL) to ensure that the learned embeddings are dataset invariant. The overall multi-dataset training approach is illustrated in Fig.~\ref{fig:flowchart}.

\subsection{Dataset-Aware Softmax Loss}
\label{sec:da}
Let us denote $\cD =\{\cD_1, \cD_2, ..., \cD_K\}$ as the training set which contains $K$ different datasets $\cD_1, \cD_2, ..., \cD_K$. We represent each training example as $(\x_i, y_i, k_{y_i})$, where $\x_i$ is the embedding vector of the $i$-th training sample, $y_i$ is the face ID label and $k_{y_i}$ presents a mapping from the face ID label $y_i$ to the dataset index $k$. The ID overlapping issue is described as two samples $\x_i, \x_j$ with the same ID $y_i=y_j$ but are from different datasets, \ie, $k_{y_i}\neq k_{y_j}$. We can naively set $y_i\neq y_j$ since the IDs from different datasets should be different; however, such ambiguity of the mislabeling issue can do harm in the training.

\begin{figure}[!t]
\centerline{\includegraphics[width=\linewidth]{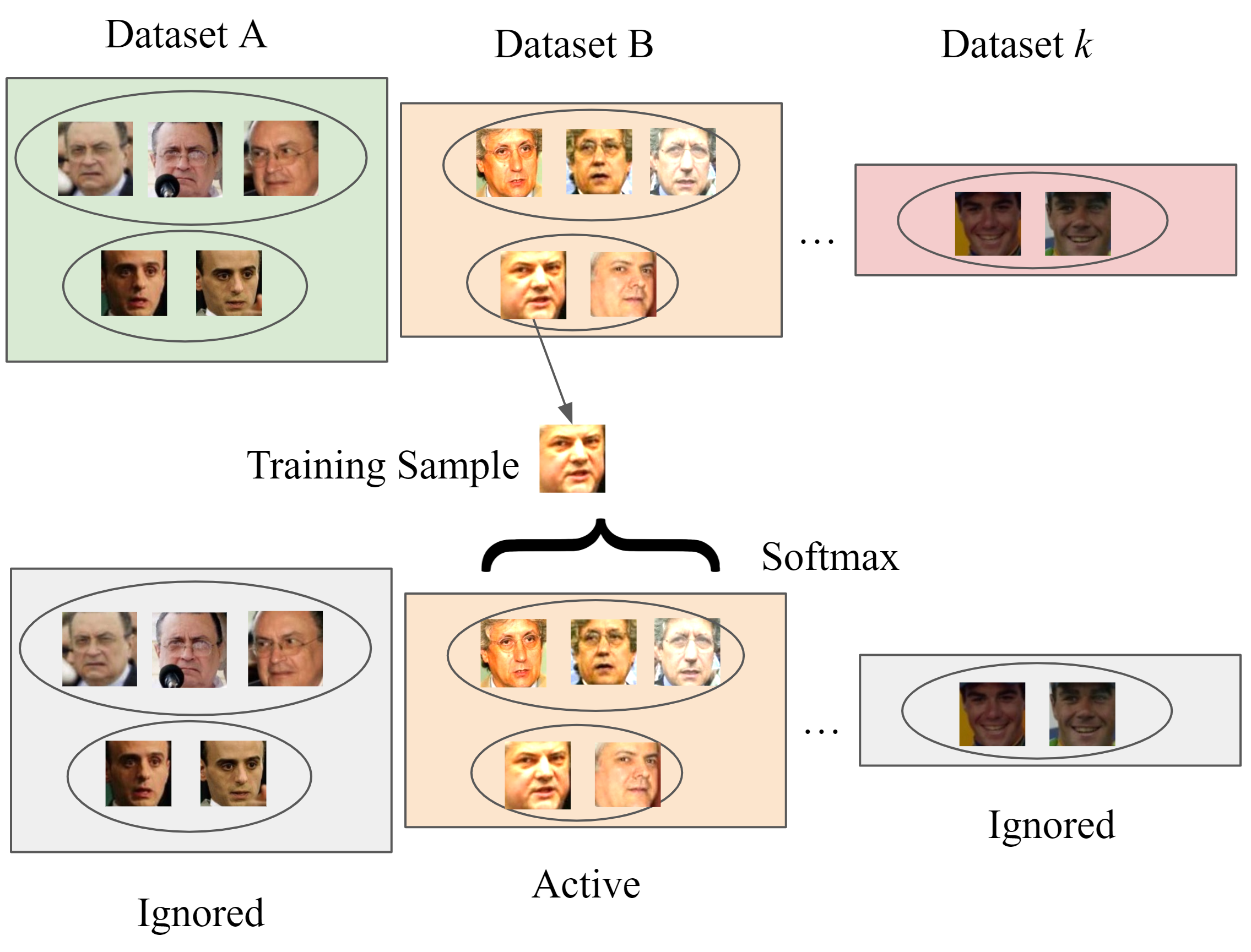}}
\caption{Example of the proposed dataset-aware softmax loss. The softmax loss is computed only within the dataset of the training samples. In this example, Dataset B is active since the training sample is selected from Dataset B, while the other datasets are ignored when computing the softmax loss.}
\label{fig:d_aware}
\end{figure}

One of the most widely used loss function in classification problems is softmax loss, which is defined as follows,
\begin{equation}
L = -\frac{1}{N}\sum_{i=1}^{N}\log\frac{e^{\bW_{y_i}^T \x_i + b_{y_i}}} {\sum_{j=1}^{C}e^{\bW_{j}^T \x_i+b_{j}}},
\label{softmax}
\end{equation}
where $\{\bW, \b\}$ are the softmax layer parameters and $C$ is the number of classes. To overcome the mislabeling issue, we define the dataset indicator to represent whether samples are from the same dataset, \ie,
\begin{equation}
\textbf{1}_{k_i=k_j}=\left\{
    \begin{array}{lr}
    1, &  \text{if} \ k_i=k_j, \\
    0, &  \text{otherwise},
    \end{array}.
\right.
\label{eq:indicator}
\end{equation}
With this, we define the dataset-aware softmax loss as follows,
\begin{equation}
L = -\frac{1}{N}\sum_{i=1}^{N}\log\frac{e^{\bW_{y_i}^T\x_i+b_{y_i}}}{e^{\bW_{y_i}^T\x_i+b_{y_i}}+\sum_{j=1, j\neq y_i}^{C}\textbf{1}_{k_j=k_{y_i}}e^{\bW_{j}^T\x_i+b_{j}}}.
\label{d_aware softmax}
\end{equation}
In other words, the softmax loss is computed within each dataset separately. As a result, the mislabeling issue can be easily solved. An example is shown in Fig.~\ref{fig:d_aware}.

Another advantage of the dataset-aware loss is that it can be combined with any variations of softmax based losses. Take ArcFace \cite{deng2019arcface} for example. The dataset-aware ArcFace can be presented as follows,
\begin{equation}
L = -\frac{1}{N}\sum_{i=1}^{N}\log\frac{e^{s\cos(\theta_{y_i}+m)}}{e^{s\cos(\theta_{y_i}+m)}+\sum_{j=1, j\neq y_i}^{C}\textbf{1}_{k_j=k_{y_i}}e^{s\cos\theta_j}},
\label{d_aware arcface}
\end{equation}
where $\theta, m$ and $s$ represent the angle, margin penalty and scale, respectively. This example shows the compatibility of the dataset-aware loss with the most advanced loss functions.

\subsection{Dataset-Invariant Learning by Domain Adaptation}
To further improve the robustness of face recognition performance across multiple domains, dataset invariant learning is crucial to ensure that the latent face embeddings are not dataset dependent. Domain adaptation with gradient reversal layers (GRL) \cite{ganin2014unsupervised} is adopted for learning the face embeddings. 

We now give more details on the domain adaptation model with GRL. For the face recognition networks, we usually have two sub-networks, a feature embedding sub-network $\x_i = G_f(I_i; \bW_f)$ and a classification sub-network $\hat{y}_i = G_y(\x_i; \bW_y)$, where $I_i$ is the input face image, $\x_i$ is the embedded feature vector, $\bW_f$ and $\bW_y$ are the network parameters. An extra domain classifier sub-network, $\hat{k}_i = G_d(\x_i; \bW_d)$, is added after the embedding network to classify which dataset the sample belongs to.
We consider two loss functions as follows,
\begin{eqnarray}
L_{cls} & =& \sum_i J_{cls}(G_y(G_f(I_i; \bW_f); \bW_y), y_i, k_i) \nonumber \\
& =& \sum_i J_{cls}(G_y(\x_i; \bW_y), y_i, k_i),
\label{cls loss}\\
L_{d} & =& \sum_i J_{d}(G_d(G_f(I_i; \bW_f); \bW_d), k_i) \nonumber \\
& =& \sum_i J_{d}(G_d(\x_i; \bW_d), k_i).
\label{dataset loss}
\end{eqnarray}
Here, $J_{cls}(\cdot)$ is the classification loss, which can be the proposed dataset-aware softmax loss and $J_{d}(\cdot)$ is the classification loss for the dataset classifier.

To learn the network parameters, we look for embeddings that minimize the classification loss as much as possible. In the meantime, a good embedding should be dataset invariant, \ie, the embedding can fool the dataset classifier so that the embedding does not correlate with the dataset. To ensure the above assumptions, the network parameters should be optimized as follows,
\begin{eqnarray}
\!\!\!\!(\hat{\bW}_f, \hat{\bW}_y) &\!\!\!\!=\!\!\!\!& \underset{\bW_f, \bW_y}{\argmin}\;\; \Big\{L_{cls}(\bW_f, \bW_y, y, k)\!
\nonumber\\
&&\;\;\;\;\;\; -\;\;\;\lambda L_{d}(\bW_f, \hat{\bW}_d, k) \Big\},
\label{optimize1}\\
\nonumber\\
\!\!\!\!\hat{\bW}_d &\!\!\!\!=\!\!\!\!& \underset{\bW_d}{\argmin}\;\; L_{d}(\hat{\bW}_f, \bW_d, k).
\label{optimize2}
\end{eqnarray}
where $\lambda$ controls the trade-off between the two objectives that shape the embeddings during learning.

\subsection{Optimization}
\label{sec:optimization}
At the beginning of the training, the dataset classifier $G_d(\cdot,\bW_d)$ is not well established. As a result, the gradient update for the feature embedding $\bW_f$ from $L_{d}(\bW_f, \hat{\bW}_d, k)$ is not quite stable and will negatively affect the discriminative ability in the classification. To further stabilize the training process, we split the optimization into two stages. The first stage is to initialize the model parameters and train the classification sub-network and dataset classifier separately as follows:
\begin{eqnarray}
(\hat{\bW}_f, \hat{\bW}_y) &=&  \underset{\bW_f, \bW_y}{\argmin} \; L_{cls}(\bW_f, \bW_y, y, k),\label{step1 optimize1}\\
\hat{\bW}_d &=& \underset{\bW_d}{\argmin}\; L_{d}(\hat{\bW}_f, \bW_d, k).\label{step1 optimize2}
\end{eqnarray}
Note that in (\ref{step1 optimize1}), the embedding sub-network $\bW_f$ is only supervised by the classification loss $L_{cls}(\bW_f, \bW_y, y, k)$  without the dataset classifier loss $L_{d}(\bW_f, \hat{\bW}_d, k)$. After each sub-network converges, we further fine-tune all the parameters based on (\ref{optimize1}) and (\ref{optimize2}) in the second stage of the training.

\section{Experiments and Results}

\subsection{Datasets and Experimental Settings}
\label{sec:data}

\noindent{\bf Datasets.} We combine 10 datasets during training,  including 14-Celebrity \cite{14Celeb}, Asian-Celeb \cite{DeepGlint}, CASIA \cite{yi2014learning}, CelebA \cite{liu2015faceattributes}, DeepGlint \cite{DeepGlint}, MS1M \cite{deng2019arcface}, PinsFace \cite{PinsFace}, 200-Celeb, VGGFace2 \cite{cao2018vggface2} and UMDFace \cite{bansal2017umdfaces}. The validation datasets include LFW \cite{huang2008labeled}, CFP-FP \cite{sengupta2016frontal} and AgeDB-30 \cite{moschoglou2017agedb}. The description of each dataset is listed as follows:
\begin{itemize}
\item 14-Celebrity \cite{14Celeb} is a small face recognition dataset for Kaggle competition, including 14 identities and 117 images.
\item Asian-Celeb \cite{DeepGlint} is a dataset contains around 94 thousand Asian celebrities with 2.8 million images.
\item CASIA \cite{yi2014learning} is created and annotated from internet faces, including more than 10 thousand identities with about 0.5 million images.
\item CelebA \cite{liu2015faceattributes} is selected from \cite{sun2014deep}, including 10 thousand identities, each of which has 20 images.
\item DeepGlint \cite{DeepGlint} is a large scale dataset with a modified combination with Asian-Celeb and MS1M datasets, including 180 thousand identities and 6.8 million images.
\item MS1M \cite{deng2019arcface} is a modified version of \cite{guo2016ms} dataset, selected from about 1 million celebrities with 6 million images.
\item PinsFace \cite{PinsFace} is collected from Pinterest and used for Kaggle face recognition competition. It includes 105 celebrities and 17534 faces.
\item 200-Celeb is self-collected face images of some Asian celebrities, including 268 identities and about 25 thousand images.
\item VGGFace2 \cite{cao2018vggface2} contains images from Google Image Search with large variations in pose, age, lighting and background. In total, it has 8.6 thousand identities and 3.1 million faces.
\item UMDFace \cite{bansal2017umdfaces} contains the images downloaded from the internet with face detection and human annotation and cleaning. In total, it has 8.3 thousand identities and 0.4 million images. 
\item LFW \cite{huang2008labeled} is the most widely used dataset for face recognition validation. It contains 5.7 thousand identities and 13 thousand faces.
\item CFP-FP \cite{sengupta2016frontal} dataset focuses on frontal to profile face verification task. It includes 500 identities and 7000 images.
\item AgeDB-30 \cite{moschoglou2017agedb} contains 16 thousand images of 568 distinct subjects. It has a large age range for each subject.
\end{itemize}
The information of each dataset is summarized in TABLE~\ref{tab1}.

\noindent{\bf Experimental Settings.} Two backbones, ResNet50 \cite{he2016deep} and MobileNetV1 \cite{howard2017mobilenets}, are used as the embedding networks, followed by two separate fully connected layers used for the classification network and the dataset classifier. The embedding dimensions for ResNet50 and MobileNet are set to be $512$ and $128$, respectively. The parameter $\lambda$ from (\ref{optimize1}) is set to $0.1$. In the training, we use a batch size of $256$ and $512$ for ResNet50 and MobileNet, respectively. The pre-trained model is adopted from \cite{deng2019arcface}. We set the initial learning rate as $0.005$, and decay it by $10$ times at steps $80000$, $140000$, and $200000$. We set the maximum steps to $240000$ for the MobileNet, while double the steps for training ResNet50. When incorporating domain adaptation in the training, we set two training stages as explained in Section \ref{sec:optimization}. We change to the second stage at the step $80000$. Four NVIDIA Tesla V100 GPUs are used in the training. 

\begin{table}[!t]
\caption{Statistics of the face datasets used in the experiments.}
\begin{center}
\begin{tabular}{|l|c|c|}
\hline
\textbf{Dataset}& \#\textbf{ID}& \#\textbf{Image}\\
\hline
14-Celebrity \cite{14Celeb}& 14& 117\\
Asian-Celeb \cite{DeepGlint}& 94.0K& 2.8M\\
CASIA \cite{yi2014learning}& 10.5K& 0.5M\\
CelebA \cite{liu2015faceattributes}& 10.2K& 0.2M\\
DeepGlint \cite{DeepGlint}& 180.9K& 6.8M \\
MS1M \cite{guo2016ms}& 85.7K& 5.8M \\
PinsFace \cite{PinsFace}& 105& 14.1K\\
200-Celeb & 268& 24.9K\\
VGGFace2 \cite{cao2018vggface2}& 8.6K& 3.1M\\
UMDFace \cite{bansal2017umdfaces}& 8.3K& 0.4M\\
\hline
LFW \cite{huang2008labeled}& 5.7K& 13,233\\
CFP-FP \cite{sengupta2016frontal}& 500& 7,000\\
AgeDB-30 \cite{moschoglou2017agedb}& 568& 16,488\\
\hline
\end{tabular}
\label{tab1}
\end{center}
\end{table}

\subsection{Compared with State-of-the-Art Methods}

Our proposed method combines the ArcFace loss with dataset-aware loss as the default setting. As a result, we show the comparison with softmax based loss in TABLE~\ref{tab2}, where the numbers in the table are the face verification accuracy. The ``Comb" in the table means combining the listed datasets from Section \ref{sec:data} in the training. From the table, we can see that the results are largely dependent on the training dataset. There is $0.3\%$ increase in accuracy if changing the CASIA dataset to the MS1M using the ArcFace loss. It is reasonable since the MS1M dataset has 8 times more IDs and is 10 times larger than the CASIA dataset. If we combine multiple datasets in the training, the accuracy on CFP-FP increases $3.1\%$ and $6.0\%$ compared with the individual CASIA and MS1M dataset, respectively. Similarly, The accuracy also increases on the AgeDB-30 dataset when combining multiple datasets in the training. This experiment shows evidence that there should be a performance gain for multi-dataset training. Such observation can be very beneficial for a practical purpose.

\begin{table}[!t]
\caption{Verification accuracy (in \%) of using different losses.}
\begin{center}
\begin{tabular}{|c|c|c|c|c|}
\hline
\textbf{Loss}& \textbf{Dataset}& \textbf{LFW}& \textbf{CFP-FP}& \textbf{AgeDB-30}\\
\hline
SphereFace& CASIA& 99.1& 94.4& 91.7\\
CosFace& CASIA& 99.5& 95.4& 94.6\\
CM (0.9, 0.4, 0.15)& CASIA& 99.5& 95.2& 94.9\\
ArcFace& CASIA& 99.5& 95.6& 95.2\\
ArcFace& MS1M& 99.8& 92.7& 97.8\\
\hline
Proposed& Comb& 99.8& 98.7& 98.2\\
\hline
\multicolumn{5}{l}{$^{\mathrm{a}}$All models are using ResNet50 for embedding.}
\end{tabular}
\label{tab2}
\end{center}
\end{table}

Apart from the comparison with ArcFace, CosFace and SphereFace, we also summarize the results compared with several state-of-the-art (SOTA) methods along with the number of training images on the LFW dataset for verification task in TABLE \ref{tab3}. The competing methods include DeepID \cite{sun2014deep}, Deep Face \cite{taigman2014deepface}, VGG Face \cite{parkhi2015deep}, FaceNet \cite{schroff2015facenet}, Baidu \cite{liu2015targeting}, Center Loss \cite{wen2016discriminative}, Range Loss \cite{zhang2017range} and Marginal Loss \cite{deng2017marginal}. Specifically, DeepID \cite{sun2014deep} extracts visual features hierarchically from local low-level to global high-level and is supervised by both identification and verification loss. Deep Face \cite{taigman2014deepface} derives the face representation by employing an explicit 3D face modeling approach. For VGG Face \cite{parkhi2015deep}, it explores the effect of using a large scale dataset in the training, while FaceNet \cite{schroff2015facenet} proposes a triplet loss for training on more than 200 million images. Baidu \cite{liu2015targeting} aggregates multi-patch information to learn the discriminative features. For Center Loss \cite{wen2016discriminative}, Range Loss \cite{zhang2017range} and Marginal Loss \cite{deng2017marginal}, loss design to generate discriminative embeddings is explored. From these methods, we can see that a large scale dataset is one important factor to achieve good performance. With our proposed approach with ResNet50 architecture, we have outperformed other SOTA methods with the help of the multi-dataset training strategy. Note that FaceNet adopts over 200 million images in the training, but this large database is not open to the public.

\begin{table}[!t]
\caption{Verification Accuracy (in \%) on LFW Compared with state-of-the-art methods.}
\begin{center}
\begin{tabular}{|c|c|c|}
\hline
\textbf{Method}&  \#\textbf{Image}& \textbf{LFW}\\
\hline
DeepID \cite{sun2014deep}& 0.2M& 99.5\\
Deep Face \cite{taigman2014deepface}& 4.4M& 97.4\\
VGG Face \cite{parkhi2015deep}& 2.6M& 99.0\\
FaceNet \cite{schroff2015facenet}& 200M& 99.6\\
Baidu \cite{liu2015targeting}& 1.3M& 99.1\\
Center Loss \cite{wen2016discriminative}& 0.7M& 99.3\\
Range Loss \cite{zhang2017range}& 5M& 99.5\\
Marginal Loss \cite{deng2017marginal}& 3.8M& 99.5\\
\hline
Proposed (ResNet50)& 19.6M& 99.8\\
\hline
\end{tabular}
\label{tab3}
\end{center}
\end{table}

\subsection{Ablation Study}
To validate the effectiveness of each component of our proposed method, we conduct ablation studies for the dataset-aware loss and dataset invariant learning and summarize the results in TABLE \ref{tab4}. In this experiment, we adopt MobileNet as the embedding network and evaluated on LFW, CFP-FP and AgeDB-30 datasets for the verification task. First, we compare the effect of different training data. Two large scale training datasets, MS1M and VGGFace2, are used for the comparison. We adopt ArcFace loss without dataset-aware and domain adaptation since only a single dataset is used in the training. The verification accuracy is shown in the first rows of TABLE \ref{tab4}. We can see that the performance on the LFW is similar with just a few percent difference on CFP-FP and AgeDB-30. This indicates that the training set indeed has a large effect on the performance. Then we also evaluate the baseline method that naively combines the ten training sets from TABLE \ref{tab1} to learn the face embedding by the default ArcFace loss. The result is shown in the 3rd row of TABLE \ref{tab4} with the method named ``Naive Comb". As expected, there is no big improvement compared with single dataset training, and there is even an accuracy drop on the LFW dataset. This is due to the label overlapping issue for multi-dataset training. Without the label cleaning techniques, the training can be sensitive and not robust with such mislabeled data. The 4th row, with the method named ``DA" (dataset-aware), shows the result with dataset-aware ArcFace loss in the training. Compared with ``Naive Comb", there is a significant improvement on all the three validation sets. This demonstrates that the label overlapping issue for multi-dataset training is automatically handled with the dataset-aware loss. The last two rows show the effectiveness incorporated with the domain adaptation approach using GRL in the training. From the results, we can see that there is a further improvement over ``DA", and obviously, ``DA+GRL" is much better than single dataset training and ``Naive Comb". As the last experiment, we also implement a ``Crossing Dropout (CD)" operation in the dataset-aware loss. Specifically, we replace (\ref{eq:indicator}) from Section \ref{sec:da} with the following modification,
\begin{equation}
\textbf{1}_{k_i=k_j,z<p}=\left\{
    \begin{array}{lc}
    1, &  \text{if} \ k_i=k_j, \text{or} \ k_i \neq k_j \ \text{and}\  z<p,\\
    0, &  \text{otherwise},
    \end{array}.
\right.
\label{eq:indicator2}
\end{equation}
where $z$ is a random variable with uniform distribution $z\sim \mathcal U(0,1)$ and $p$ is a pre-defined probability, set as $0.0001$ in the experiment. Different from the original dataset-aware loss, this modification only includes a few classes from other datasets for each training sample. Setting $p=1$ will degrade to the original softmax loss and $p=0$ is the proposed dataset-aware loss. With a small value $p$, the randomly selected classes are not likely to be the overlapping class of the training sample. Meanwhile, it can also help the network learn discrimination across different datasets. Interestingly, we find that setting $p$ to be a small value of $0.0001$, there is an improvement on CFP-FP and AgeDB-30 datasets and also comparable on the LFW dataset.

\begin{table}[!t]
\caption{Ablation Study using MobileNet. There numbers are accuracies (in \%).}
\begin{center}
\begin{tabular}{|c|c|c|c|c|}
\hline
\textbf{Method}& \textbf{Dataset}& \textbf{LFW}& \textbf{CFP-FP}& \textbf{AgeDB-30}\\
\hline
ArcFace& MS1M& 99.5& 88.9& 95.9\\
ArcFace& VGGFace2& 99.5& 94.2& 93.6\\
Naive Comb& Comb& 99.1& 95.0& 94.8\\
DA& Comb& 99.5& 95.4& 95.7\\
DA+GRL & Comb& 99.7& 96.0& 96.0\\
DA+GRL+CD& Comb& 99.6& 96.3& 96.2\\
\hline
\end{tabular}
\label{tab4}
\end{center}
\end{table}

\section{Conclusion}
In this paper, a dataset-aware loss with dataset invariant learning approach is presented for face recognition to address multi-dataset training issues, including ID overlapping issue and domain distribution mismatches. From the experiments, the proposed dataset-aware loss outperforms the single dataset training and naive combining  strategy. Dataset-invariant learning with domain adaptation can further improve the verification accuracy.

\bibliographystyle{IEEEtran}
\bibliography{ref}

\end{document}

%% file: definitions.tex
\usepackage{amsmath}
\usepackage{etoolbox}
\usepackage{bm}
\usepackage{amsthm}
\usepackage{mathtools}
\usepackage{algorithm}
\usepackage{algpseudocode}
\usepackage{xspace}

\newcommand{\MyMapTemplateNoPrefix}[3]{\expandafter#1\csname#3\endcsname{#2{#3}}}
\forcsvlist{\MyMapTemplateNoPrefix {\def} {\mathbf} } {0,1,a,b,c,d,e, f, g, h, i, j, k, l, m, n, o, p, q, r, u, v, w, x, y, z} 

\newcommand{\MyMapTemplatePrefix}[4]{\expandafter#1\csname#3#4\endcsname{#2{#4}}} 
\forcsvlist{\MyMapTemplatePrefix {\def} {\mathcal}{c}} {A,B,C,D,E,F,G,H,I,J,K,L,M,N,O,P,Q,R,S,T,U,V,W,X,Y,Z}  
\forcsvlist{\MyMapTemplatePrefix {\def} {\mathbf} {b}} {A,B,C,D,E,F,G,H,I,J,K,L,M,N,O,P,Q,R,S,T,U,V,W,X,Y,Z} 
\forcsvlist{\MyMapTemplatePrefix {\DeclareMathOperator} {} {} } {trace, argmax, argmin, CE, BCE, Ent, MMD, WCD, MCD}

\def\ie{\emph{i.e.}\@\xspace}